\documentclass[letterpaper, 10 pt, conference]{ieeeconf}

\IEEEoverridecommandlockouts
\usepackage{cite}
\usepackage{amsmath,amssymb,amsfonts}
\usepackage{graphicx}
\usepackage{textcomp}
\usepackage{xcolor}
\usepackage{colortbl}

\newcommand{\zerocell}{\cellcolor{red!15}0}          %
\newcommand{\initzero}[1]{\cellcolor{black!8}#1}     %
\newcommand{\best}[1]{\textbf{#1}}                   %
\usepackage{booktabs}
\usepackage[hidelinks]{hyperref}
\usepackage{subcaption}
\usepackage[utf8]{inputenc}

\makeatletter
\def\bstctlcite{\@ifnextchar[{\@bstctlcite}{\@bstctlcite[@auxout]}}
\def\@bstctlcite[#1]#2{\@bsphack
  \@for\@citeb:=#2\do{%
    \edef\@citeb{\expandafter\@firstofone\@citeb}%
    \if@filesw\immediate\write\csname #1\endcsname{\string\citation{\@citeb}}\fi}%
  \@esphack}
\makeatother

\begin{document}
\bstctlcite{BSTcontrol}

\title{\LARGE \bf
SynthDemo-RL: Breaking the Zero-Reward Barrier \\ in VLA Adaptation with LLM-Guided Synthetic Demonstrations
}

\author{Hiroaki Kingetsu$^{1}$, Hiroaki Kurihara$^{1}$, Kaoru Yokoo$^{1}$, Kenji Fukumizu$^{2,1}$, and Manohar Kaul$^{1}$%
\thanks{\raggedright $^{1}$Fujitsu Limited, Kawasaki, Japan.}%
\thanks{\raggedright $^{2}$The Institute of Statistical Mathematics, Tokyo, Japan.}%
\thanks{\raggedright Correspondence: {\tt\small h.kingetsu@fujitsu.com}}%
}

\maketitle
\thispagestyle{empty}
\pagestyle{empty}

\begin{figure*}[t]
    \centering
    \includegraphics[width=\textwidth]{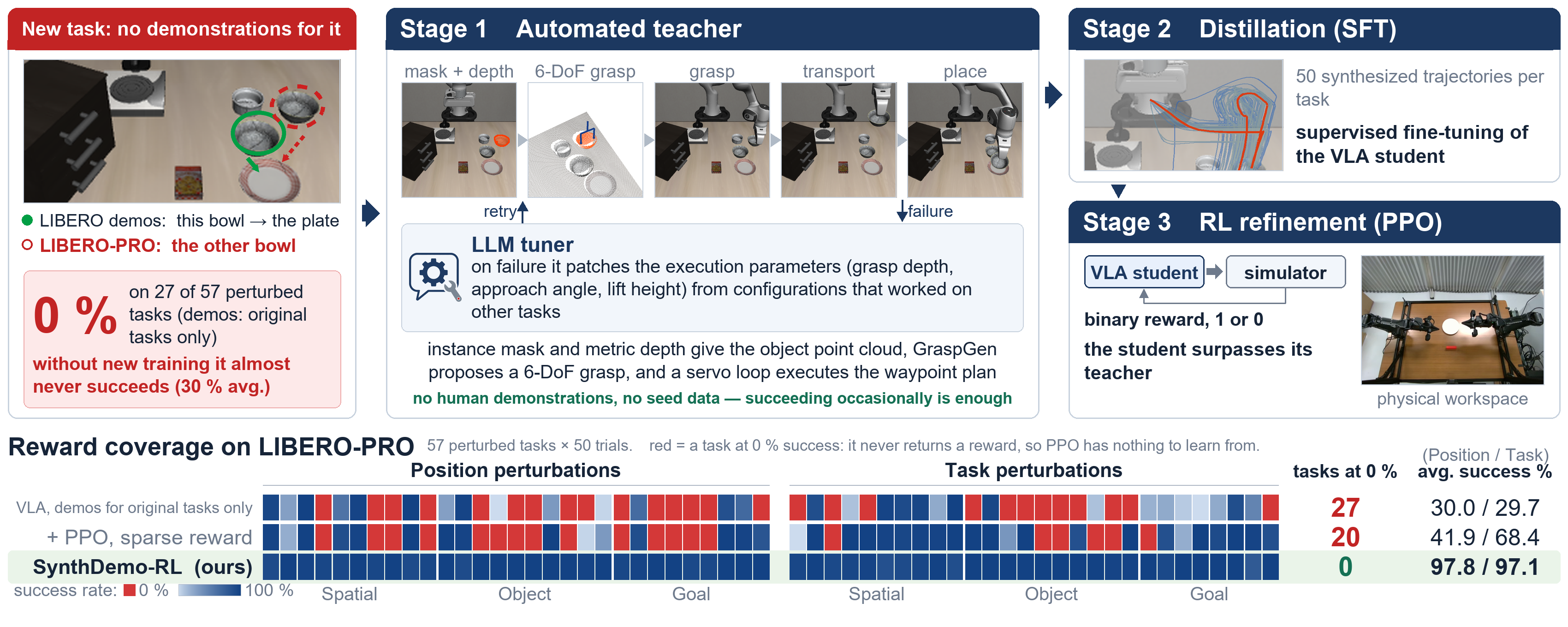}
    \caption{
        \textbf{SynthDemo-RL: VLA task adaptation without new human demonstrations.}
        Human demonstrations exist only for the original task (green marker).
        LIBERO-PRO asks for the other bowl (red), and a VLA fine-tuned on the original tasks has no observed successes on 27 of the 57 perturbed tasks under the evaluation protocol.
        An automated teacher solves the perturbed task from simulator-privileged state, SFT distills its successful trajectories into the VLA student, and PPO refines the student past its teacher.
        \textbf{Bottom:} per-task success (Table~\ref{tab:pertask}).
        Direct PPO leaves 20 of the 57 tasks with no observed success after adaptation (17 unrescued of the 27, plus 3 driven to 0\%), while ours leaves none.}
    \label{fig:pipeline}
\end{figure*}

\begin{abstract}
Fine-tuning Vision-Language-Action (VLA) models commonly relies on human teleoperation demonstrations, while reinforcement learning (RL) with sparse binary rewards faces an exploration challenge when successful trajectories are rarely sampled.
We propose \textbf{SynthDemo-RL}, a teacher-student framework in which an automated teacher converts simulator-privileged state into successful manipulation trajectories, a VLA student is distilled from them by supervised fine-tuning (SFT), and PPO with binary task-success rewards refines the student.
We study \emph{reward coverage}, the fraction of tasks for which at least one success is observed under the fixed evaluation protocol, as a complement to the average success rate.
On LIBERO-PRO, a public benchmark of perturbed LIBERO tasks for which no demonstrations exist, 27 of 57 scored tasks are at exactly 0\% success for a $\pi_{0.5}$ policy fine-tuned on the original LIBERO tasks.
Direct PPO from this policy, under the same PPO recipe and the same RL compute as SynthDemo-RL's refinement stage, rescues 10 of these 27 tasks and leaves 17 at 0\%.
SynthDemo-RL, with 50 synthesized trajectories per task and no new human demonstrations, rescues all 27 and reaches average success rates of 97.8\% and 97.1\% on the Position and Task axes of LIBERO-PRO, respectively.
On standard LIBERO, the same pipeline reaches 96.0\% with no human demonstrations, within 1.7 points of $\pi_{0.5}$ trained on 50 human demonstrations per task.
We further validate the pipeline on RoboTwin~2.0 and verify that trajectories from a policy trained in a MuJoCo twin execute open-loop on a physical robot.
\end{abstract}

\section{Introduction}
\label{sec:introduction}

Vision-Language-Action (VLA) models~\cite{openvla} have shown strong generalization in robotic manipulation through internet-scale vision-language pretraining.
Adapting them to specific downstream tasks, however, commonly relies on imitation learning through supervised fine-tuning (SFT) on human teleoperation demonstrations, which fundamentally limits the scalability of VLA-based systems across tasks and environments.

Recent work shows that reinforcement learning (RL) can lift VLA policies far above their SFT baseline, even from a single demonstration~\cite{simplevla-rl,vla-rl,rlinf-vla}, but it still assumes human-collected demonstrations.
When successful trajectories are rarely sampled, RL with sparse binary rewards may struggle to discover rewarding behavior.

This zero-reward regime is our starting point. Demonstrations are the classical way out~\cite{dapg}, and unlike reward shaping they need no per-task reward design~\cite{eureka,robogen}. If an automated procedure can solve a task even occasionally, its successful trajectories can be distilled into the policy as an initialization for sparse-reward RL. We therefore distinguish \emph{average success rate} from \emph{reward coverage}, the fraction of tasks with at least one observed success. Average success measures how well a policy performs; coverage measures on how many tasks RL has anything to build on. A high average can hide tasks with no observed success, and those are exactly where sparse-reward RL fails.

We propose \textbf{SynthDemo-RL}, a three-stage framework for training VLA manipulation policies without new human demonstrations (Fig.~\ref{fig:pipeline}).
An automated teacher with simulator-privileged state collects successful trajectories. A pretrained grasp predictor proposes grasps and a scripted controller executes a waypoint sequence (approach, grasp, lift, place). When an attempt fails, an LLM inspects the failure log and images. It then revises the waypoint sequence and execution parameters such as contact depth and lateral retreat. The teacher prioritizes coverage over optimality and keeps trying until every task yields successful trajectories.
The teacher is too slow to run online, so SFT distills its trajectories into a VLA student. PPO with binary task-success rewards then refines the student past its teacher.
We demonstrate that synthetic demonstrations can initialize sparse-reward RL at VLA scale without collecting new human demonstrations for the target tasks.
SFT achieves nonzero success on every task, and RL substantially improves the resulting policies.

In Sec.~\ref{sec:exp:novel} we study human-demonstration-free adaptation on LIBERO-PRO~\cite{liberopro}, a public benchmark that systematically perturbs LIBERO tasks (e.g., moving initial object placements or redefining task goals); such perturbations severely degrade policies fine-tuned on human demonstrations, and no demonstrations exist for the perturbed variants.

Our contributions are:
\begin{itemize}
    \item SynthDemo-RL, a pipeline that combines an automated teacher with VLA distillation and sparse-reward PPO. The teacher converts simulator-privileged state into successful manipulation trajectories with no human demonstrations or task-specific training.
    \item Reward coverage, the fraction of tasks with at least one observed success, as a diagnostic complementary to average success. On perturbed LIBERO-PRO, distilling synthesized trajectories rescues all 27 tasks with no observed success before adaptation, while the PPO-matched sparse-reward control leaves most of them at 0\%.
    Among tasks with nonzero SFT success, RL lifts even weak initializations to high final success.
    \item An analysis of synthetic SFT limitations. Synthetic trajectories differ systematically from human trajectories in motion statistics and frozen-base flow-matching loss.
    SFT on regenerated demonstrations with reduced measured differences remains well below human-demonstration SFT.
\end{itemize}

\section{Related Work}
\label{sec:related}

\subsection{Reinforcement Learning for VLA Fine-Tuning}

SimpleVLA-RL~\cite{simplevla-rl}, VLA-RL~\cite{vla-rl}, RLinf-VLA~\cite{rlinf-vla}, and VLA-RFT~\cite{vla-rft} refine VLA policies with online RL.
They differ in policy interface, infrastructure, and whether rollouts use a simulator or a learned world model, but all assume human-collected SFT data for initialization.
The binary task-success reward is the common choice in this line~\cite{simplevla-rl,rlinf-vla,pirl}.
Our contribution is to show that synthesized demonstrations can initialize sparse-reward RL at VLA scale on target task distributions without human demonstrations.

\subsection{Automated Data Generation and Learning without Human Demonstrations}

MimicGen~\cite{mimicgen} and RoboCasa~\cite{robocasa} expand demonstration datasets through retargeting but require source demonstrations of the target behavior. Our teacher generates target-task demonstrations without such sources.
Ha~et~al.~\cite{scalingup} generate task plans with an LLM, solve them with a sampling-based planner under success verification, and distill into a language-conditioned policy.
SynthDemo-RL is closest to this line in structure.
We extend this synthesis-and-distillation approach to VLA adaptation, using the distilled policy as the initialization for sparse-reward RL.
Among approaches that aim to remove demonstrations entirely, ReWiND~\cite{rewind} still needs initial demonstrations for its reward model. Self-Improving VLA~\cite{pld} needs an initial policy that already succeeds, and GigaBrain~\cite{gigabrain} needs a proprietary world model.
LLMs have served as automated tuning agents for rewards and task specifications~\cite{eureka,robogen}.
Our LLM tuner instead revises waypoint sequences and execution parameters inside the teacher to raise data-collection success.

\subsection{Teacher-Student Policy Distillation}

Distilling non-learned or privileged teachers into deployable students is well established: Neural~MP~\cite{neural-mp} distills a sampling-based motion planner into a policy that surpasses its teacher, and privileged-teacher training is standard in locomotion~\cite{privileged-teacher}.
RPD~\cite{rpd} distills a VLA teacher into an RL student but queries the teacher online at every step, and the teacher itself was trained on demonstrations.
Our teacher requires no task-specific training, and distillation is offline: the teacher generates data once and is discarded.

\section{Method: SynthDemo-RL}
\label{sec:method}

\subsection{Overview}
\label{sec:method:overview}
The goal is to learn a VLA policy $\pi_\theta$ that maximizes task success rate in a simulation environment without collecting new human demonstrations for the target tasks.
The teacher generates $\mathcal{D} = \{(\tau_i, r_i)\}_{i=1}^N$, where $N$ is the number of recorded trials, $\tau_i$ is an observation-action trajectory, and $r_i \in \{0,1\}$ is the outcome of the simulator's success checker.
The successful trajectories $\mathcal{D}^+ = \{\tau_i \mid r_i = 1\}$ are used to train $\pi_\theta^{\text{SFT}}$.
PPO with binary task-success rewards then refines it into $\pi_\theta^{\text{RL}}$. 

\subsection{Automated Teacher Pipeline}
\label{sec:method:teacher}

The teacher takes a language instruction and access to the simulator state (ground-truth object poses, metric depth, instance segmentation, camera parameters, and the end-effector pose).
It executes end-effector actions and records the resulting observation–action trajectory $\tau$.
It is a scripted executor around two learned components, a pretrained grasp generator and an LLM tuner.

\subsubsection{Goal Specification and Object Grounding}
The task definition identifies the manipulated object and its goal region, and the simulator provides their ground-truth poses.
The object's instance mask and metric depth are back-projected through the camera parameters into a 3D point cloud, which is the geometric input to grasping.

\subsubsection{Grasp and Placement}
A neural grasp predictor (GraspGen~\cite{graspgen}) proposes 6-DoF grasps on the object point cloud.
Candidates are filtered to the object's axis-aligned bounding box and post-processed to use a top-down approach by default with a tunable contact depth (Sec.~\ref{sec:method:teacher:tuning}).
The placement pose is computed from the goal object's ground-truth top-surface position with a clearance offset.

\subsubsection{Closed-Loop Execution}
The LLM tuner (GPT-5.5~\cite{gpt55}) configures the waypoint sequence within each manipulation primitive, including its stages and their order, as well as execution parameters.
A pick-and-place plan may include pre-grasp, grasp, lift, pre-place, place, and retreat, while a drawer-opening plan may use a contact-sweep-retreat sequence.
A proportional servo tracks these waypoints with per-step position correction, which matters during contact-rich grasp phases where open-loop execution diverges.

\subsubsection{LLM-Guided Tuning}
\label{sec:method:teacher:tuning}
After failed attempts, the LLM tuner revises the waypoint sequence and execution parameters, including contact depth, approach angle, lift and lateral retreat, and placement offset.
It receives the failure stage and reason from the execution log, observation images of the failed attempt, a \emph{success library} of configurations that worked on other tasks (ranked by relevance), and a \emph{failure library} of configurations already tried on this task.
From these it proposes a revised configuration for the next attempt.
The success library transfers configurations across geometrically similar tasks; the failure library prevents revisiting dead ends.

\subsubsection{Data Collection and Filtering}
\label{sec:method:teacher:data}
During execution the teacher records multi-camera observations (third-person and wrist), 7-DoF actions, and proprioceptive state.
Post-execution filtering retains only successful episodes and removes no-op transitions, and the result is exported to LeRobot format.

\subsection{Offline Policy Distillation}
\label{sec:method:sft}

The teacher needs ${\sim}1.7$ minutes per episode (measured on LIBERO-Object) for grasp inference and closed-loop planning, incompatible with 10--20\,Hz control, so we distill its behavior into a VLA policy that predicts action chunks directly from observations.
The student is $\pi_{0.5}$~\cite{pi05}, a flow-matching VLA that predicts continuous action chunks from third-person and wrist images, proprioceptive state, and the instruction.
We fine-tune on the successful teacher trajectories $\mathcal{D}^+$, starting from the released $\pi_{0.5}$ base checkpoint for standard LIBERO and, for LIBERO-PRO adaptation, from $\pi_{0.5}$-LIBERO, our own human-demonstration SFT of that base (row (d) of Table~\ref{tab:libero_main}).
Both settings train one policy per suite (on LIBERO-PRO, one per suite and perturbation axis, six conditions in total) on 50 successful episodes per task. LIBERO-PRO adaptation uses only trajectories synthesized on each perturbed task, with no original-LIBERO replay.
Training follows the standard flow-matching objective and the openpi recipe: 30{,}000 steps, batch size 32, peak learning rate $5\times10^{-5}$, 8 MI300X GPUs.
Sec.~\ref{sec:exp:whysft} examines the SFT gap and the effect of modifying the teacher's motion characteristics.

\subsection{RL Refinement}
\label{sec:method:rl}

We apply PPO~\cite{ppo} through the RLinf framework~\cite{rlinf-vla} to the SFT-initialized student, with the binary task-success reward and no shaping.
This keeps the optimization target identical to the evaluation metric and needs no per-task reward engineering.
To obtain a tractable likelihood for PPO, we use Flow-SDE~\cite{pirl}: Gaussian noise at each denoising step yields transitions with closed-form densities, whose log-densities sum to the sampled denoising-path log-probability used in the PPO ratio.
Each MDP step executes one action chunk, the sparse reward arrives at episode termination, and GAE~\cite{gae} operates over chunk steps using a value head on the $\pi_{0.5}$ backbone, with no separate critic model.
Each iteration collects eight rounds of rollouts, each running 64 parallel environments for 240 chunk steps.
We use three denoising steps with exploration noise 0.5 and five-action chunks of 7-DoF actions.
Each PPO update runs one epoch over a global batch of 2048 chunk steps (micro batch 128) with $\gamma=0.99$, GAE $\lambda=0.95$, clip 0.2, dual clip 3.0, value clip 0.2, and no KL penalty or entropy bonus.
The actor and value head use learning rates of $5\times10^{-6}$ and $1\times10^{-4}$ with gradient clip 1.0.
RL serves two purposes: it recovers from the biases of the teacher's scripted execution by visiting states the teacher never produced, and it lets the student surpass the teacher's own reliability, as also observed for distilled planners~\cite{neural-mp,rpd}.

\section{Experiments}
\label{sec:experiments}

\subsection{Experimental Setup}
\label{sec:exp:setup}

\subsubsection{Benchmarks}
\label{sec:exp:benchmarks}

\textbf{LIBERO}~\cite{libero} is a single-arm manipulation benchmark (Franka Panda, MuJoCo).
We evaluate on the three suites targeting single-stage manipulation: \textbf{LIBERO-Spatial} (10 tasks, spatial relations), \textbf{LIBERO-Object} (10 tasks, object identity), and \textbf{LIBERO-Goal} (9 tasks, goal variation).%
\footnote{We restrict evaluation to single-stage manipulation, which excludes LIBERO-Long and LIBERO-Goal Task~3. Task~3 chains drawer opening with a placement; the teacher supports both primitives individually but does not implement their sequencing (Sec.~\ref{sec:limitations}). It is excluded from all of our runs, in training and scoring.}
\textbf{LIBERO-PRO}~\cite{liberopro} applies systematic perturbations to these suites; we use the two axes under which fine-tuned VLAs degrade most: \textbf{Position}, which reassigns objects to alternative placement regions while keeping goals and instructions unchanged, and \textbf{Task}, which rewrites goal predicates and instructions while keeping the object set (e.g., ``open the \underline{middle} drawer'' $\to$ ``open the \underline{bottom} drawer'').%
\footnote{We exclude Task-Goal Task~0 from all aggregates of our runs. The unchanged layout leaves a plate and a bowl in the bottom drawer's swept path, so opening it shoves them aside (${\sim}9$ and 15\,cm in all 50 evaluation states), and the evaluated rollouts satisfy the success checker despite this displacement. The perturbation thus changes the required physical behavior, not only the instruction as the Task axis intends, and we judge the instance ill-posed. The Task axis scores 28 tasks.}
We use the officially distributed perturbed task definitions and evaluation initial states, frozen with content hashes before any experiment.
Our Position-Goal and Task-Goal evaluations score 9 and 8 tasks, respectively, and each task is evaluated over 50 trials.
\textbf{RoboTwin~2.0}~\cite{robotwin2} is a dual-arm benchmark in SAPIEN.
We use its single-arm Agilex Piper configuration on 4 tasks as a cross-simulator, cross-embodiment test (bimanual coordination is out of scope).

\subsubsection{Methods Compared}

We compare \textbf{(a) SynthDemo-Teacher} (the automated teacher of Sec.~\ref{sec:method:teacher}), \textbf{(b) SynthDemo-SFT} (teacher data plus SFT of the $\pi_{0.5}$ student), and \textbf{(c) SynthDemo-RL} (ours: teacher data plus SFT plus RL).
Methods (a)--(c) use \textbf{zero human demonstrations}.
On LIBERO, we further compare \textbf{(d) $\pi_{0.5}$ (human)} (the same student fine-tuned by us on 50 human teleoperation demonstrations per task, one policy per suite) and \textbf{(d') $\pi_{0.5}$ (human) $+$ PPO} (row (d) refined under the same RL recipe as (c)).
Rows (d, d') are run by us under our protocol.
Row (d) is trained per suite from the released $\pi_{0.5}$ base on the 29 scored tasks and is distinct from the publicly released $\pi_{0.5}$-LIBERO checkpoint, a single policy trained jointly on all four suites including LIBERO-Long and then evaluated suite by suite. The two are not comparable.
On RoboTwin only, we quote the published \textbf{(e) OpenVLA-OFT} and \textbf{(e') SimpleVLA-RL} results~\cite{simplevla-rl}, which use a different policy backbone. Row (e) is SFT on the benchmark's generated expert demonstrations, and row (e') adds RL.

\subsubsection{Evaluation Protocol and Initial-State Provenance}
\label{sec:exp:protocol}

Success is determined by the simulator's built-in success checker.
We report per-suite averages of per-task success rates and, on LIBERO-PRO, \emph{reward coverage}, the fraction of tasks with at least one observed success in the reported evaluation trials.
Our LIBERO-PRO results use three SFT training seeds, and the RL stage starts from the seed
whose average success rate is the median across the six suite-axis conditions.
For the per-task results, the coverage-sensitivity analysis, and the correlation between
initialization and final success, SynthDemo-SFT is evaluated at the checkpoints used to
initialize RL.
Each seed is evaluated over 50 trials per task, for which the binomial standard error is
about 7 points for a task at 50\% success and 2.4 points at 97\%, and about 2.2 and 0.8
points for a 10-task suite mean.

On \textbf{standard LIBERO}, synthesis uses initial states disjoint from the benchmark's fixed array of 50 evaluation initial states per task, and the human demonstrations are likewise a disjoint sample from the same distribution (no synthesis or human-demonstration initial state coincides with an evaluation state on any of the three suites).
Rows (a)--(d) therefore share initial-state provenance, and Sec.~\ref{sec:exp:libero} compares them at matched policy, training budget, and episode count.
Both RL arms draw rollouts from the official initial states, as is standard in LIBERO RL work.
On \textbf{LIBERO-PRO}, training initial states for synthesis and RL come from seeded environment resets, with an automated check that rejects any state coinciding with the officially distributed set, which is reserved for evaluation.

\subsection{Standard LIBERO}
\label{sec:exp:libero}

Table~\ref{tab:libero_main} asks whether the pipeline can reach human-demonstration-level performance without any human demonstrations; rows (b) and (d) differ only in the source of the 50 SFT episodes per task (Sec.~\ref{sec:exp:protocol}).

\begin{table}[t]
    \centering\small
    \caption{Success rates (\%) on LIBERO (50 trials/task). Row (a): teacher single-attempt success during data generation. Retained demonstrations are 100\% successful by construction. Rows (b, c): fixed final checkpoints under the prescribed training schedules. Rows (d, d'): our human-demonstration reference (50 demos/task). Goal is the 9-task subset of Sec.~\ref{sec:exp:benchmarks}.}
    \label{tab:libero_main}
    \resizebox{\columnwidth}{!}{%
    \begin{tabular}{@{}lcccc@{}}
        \toprule
        Method & Spatial & Object & Goal & Avg. \\
        \midrule
        \multicolumn{5}{@{}l}{\textit{Zero human demonstrations (ours)}} \\
        (a) SynthDemo-Teacher (1-attempt)     & 79.9 & 98.0 & 43.0 & 73.6 \\
        (b) SynthDemo-SFT    & 55.3 & 69.8 & 45.1 & 56.7 \\
        (c) \textbf{SynthDemo-RL (ours)} & \textbf{96.2} & \textbf{98.6} & \textbf{93.3} & \textbf{96.0} \\
        \midrule
        \multicolumn{5}{@{}l}{\textit{Human-demonstration reference}} \\
        (d) $\pi_{0.5}$ + human-demo SFT & 97.8 & 99.0 & 96.2 & 97.7 \\
        (d') $\pi_{0.5}$ + human-demo SFT + PPO & 95.8 & 99.0 & 97.3 & 97.4 \\
        \bottomrule
    \end{tabular}}
\end{table}

SynthDemo-RL (c) reaches 96.0\% without human demonstrations, within 1.7 points of both human-demonstration references.
Two observations set up the rest of the paper.
First, distillation alone remains well below human-demonstration SFT at matched policy, budget, and episode count (Sec.~\ref{sec:exp:whysft}).
Second, RL rather than distillation closes the gap.
RL gains about 40 points over SFT and surpasses the teacher's single-attempt success on every suite, most markedly on Goal (93.3 vs 43.0).
Generating one retained demonstration costs 0.4--3.4 minutes depending on the suite.

\subsection{Adaptation to Perturbed Tasks without New Human Demonstrations}
\label{sec:exp:novel}
Using LIBERO-PRO, we test whether a VLA fine-tuned on the original tasks can adapt to shifted task distributions without new human demonstrations.

We adapt $\pi_{0.5}$-LIBERO, i.e.\ our per-suite row (d) of Table~\ref{tab:libero_main}, with trajectories synthesized directly on each perturbed task.
The teacher retains 50 successful trajectories per task, and SFT uses these only (no original-LIBERO replay, an assumption tested below).
As the sparse-reward control, direct PPO starts from the same policy and uses the same PPO recipe as SynthDemo-RL.
The direct-PPO control and SynthDemo-RL use identical PPO iteration counts and environment interactions within each suite-axis condition, averaging 188 iterations (23.1 million chunk steps) across the six conditions.
The only difference is the target-task synthesis and SFT that precede RL in SynthDemo-RL.
For Task perturbations the rewritten instruction is supplied during training and evaluation.%
\footnote{As released, the official evaluation feeds the pre-perturbation instruction while scoring the perturbed goal, a mismatch also reported independently in the benchmark's issue tracker. It measures blind instruction-following, so our Task-axis numbers are not comparable to the benchmark's released numbers; the quoted rows of Table~\ref{tab:libero_pro} also feed the rewritten instruction (note $\ddagger$) and are unaffected.}

Table~\ref{tab:libero_pro} reports the results.
Before adaptation, 27 of the 57 scored perturbed tasks are at exactly 0\%.

The teacher yields at least one successful trajectory on every one of the 57 tasks.
On average the tuner produced 3.3 further distinct execution configurations per task beyond the initial one (SD 6.3 over the 57 tasks).
Without LLM revision, rerunning each suite's frozen initial configuration (25 attempts per task) leaves 12 of the 57 tasks with no successful trajectory and gives task-mean success of 54.4\% (Position) and 47.6\% (Task).
Table~\ref{tab:libero_pro} reports teacher success rates using each task's final tuned configuration (65.6\% and 71.0\%); attempt counts vary per task because generation stops once 50 successful trajectories are retained.
The 12 tasks involve articulated drawers and a stove knob, placements onto a stove, rack, or cabinet top, or flat and elongated containers.
The tuned configurations rescue these tasks by switching to rim or pinch grasps and adjusting grasp depth and lateral retreat.

\begin{table*}[t]
\centering\footnotesize
\caption{Task success rates (\%) and reward coverage on \textbf{LIBERO-PRO} (50 trials/task for our runs). For evaluated policies, ``Coverage''\ counts tasks per axis with at least one observed success in the reported trials (teacher: at least one successful synthesis; success rates use each task's final tuned configuration). SFT success rates are averaged over three training seeds, and $\pm$ denotes the across-seed standard deviation of the axis average. Every seed achieves at least one success on all 57 tasks in 50 evaluation trials per task.}
\label{tab:libero_pro}
\begin{tabular}{@{}lcccccccccc@{}}
\toprule
& \multicolumn{5}{c}{\textbf{Position}} & \multicolumn{5}{c}{\textbf{Task}} \\
\cmidrule(lr){2-6} \cmidrule(lr){7-11}
Method & Spatial & Object & Goal$^\dagger$ & Avg. & Coverage & Spatial & Object & Goal$^\dagger$ & Avg. & Coverage \\
\midrule
\multicolumn{11}{l}{\textit{Before adaptation}} \\
~~$\pi_{0.5}$-LIBERO & 48.4 & 16.8 & 24.7 & 30.0 & 14/29 & 51.2 & 10.8 & 27.0 & 29.7 & 16/28 \\
\midrule
\multicolumn{11}{l}{\textit{Existing approaches (quoted)}$^\ddagger$} \\
~~SPARK~\cite{spark} & 56.0 & 43.4 & 40.0 & 46.5 & -- & 72.4 & 36.4 & 14.0 & 40.9 & -- \\
~~Pigey~\cite{pigey} & 66.0 & 54.0 & 44.0 & 54.7 & -- & 80.0 & 54.0 & 22.0 & 52.0 & -- \\
~~CounterAlign~\cite{counteralign} & 60.0 & 51.0 & 41.0 & 50.7 & -- & 63.0 & 26.0 & 46.0 & 45.0 & -- \\
\midrule
\multicolumn{11}{l}{\textit{Sparse-reward control, matched RL compute}} \\
~~$\pi_{0.5}$-LIBERO $+$ PPO & 50.8 & 33.6 & 41.3 & 41.9 & 15/29 & 79.2 & 52.4 & 73.5 & 68.4 & 22/28 \\
\midrule
\multicolumn{11}{l}{\textit{SynthDemo adaptation}} \\
~~SynthDemo-Teacher (1-attempt) & 60.0 & 68.0 & 68.9 & 65.6 & 29/29 & 60.1 & 70.6 & 82.4 & 71.0 & 28/28 \\
~~SynthDemo-SFT & 57.3 & 56.5 & 54.8 & 56.2$\pm$1.3 & \textbf{29/29} & 53.3 & 66.3 & 42.7 & 54.1$\pm$1.9 & \textbf{28/28} \\
~~\textbf{SynthDemo-RL (ours)} & \textbf{96.4} & \textbf{99.0} & \textbf{98.0} & \textbf{97.8} & \textbf{29/29} & \textbf{95.0} & \textbf{99.6} & \textbf{96.8} & \textbf{97.1} & \textbf{28/28} \\
\bottomrule
\end{tabular}
\par\vspace{2pt}
\begin{minipage}{\textwidth}\footnotesize
$^\dagger$For our runs, Goal scores 9 tasks on Position (Task~3 excluded) and 8 on Task (Tasks~0 and 3 excluded, Sec.~\ref{sec:exp:benchmarks}); quoted rows score 10.
$^\ddagger$Quoted results follow their source protocols (own evaluation harnesses; 10, 50, and 100 trials per task for Pigey, SPARK, and CounterAlign; the rewritten instruction on the Task axis, as in our runs) and are contextual references, not controlled comparisons.
\end{minipage}
\end{table*}

SynthDemo-RL exceeds direct PPO by 28.7/55.9 points on Task/Position.
Sparse-reward exploration does find some tasks on its own: over its RL budget the control rescues 10 of the 27 tasks with no observed success before adaptation.
But 17 of the 27 still have no observed successes after control training (13 of 15 on Position, 4 of 12 on Task).
SynthDemo-SFT, by contrast, makes every one of the 27 nonzero before any RL. This coverage result holds across all three SFT seeds, each achieving at least one success on every one of the 57 tasks in 50 evaluation trials per task.
Coverage depends on the trial budget. Had only 10 of the 50 recorded trials per task been run, the expected number of covered tasks (averaged over all 10-trial subsets) would be 26.6 for $\pi_{0.5}$-LIBERO, 35.4 for the control, 53.6 for SynthDemo-SFT, and 57 for SynthDemo-RL, out of 57.
Even when coverage requires at least five successes in 50 trials, SynthDemo-SFT covers 53 of 57 tasks and SynthDemo-RL covers all 57.

For context, Table~\ref{tab:libero_pro} also quotes published LIBERO-PRO results. SPARK and Pigey use LLM-based inference-time planning or orchestration, whereas CounterAlign applies counterfactual offline RL to LIBERO demonstrations.

Across the 57 tasks and both RL arms (114 task--arm pairs, where the initialization is $\pi_{0.5}$-LIBERO for the control and SynthDemo-SFT for ours), whether the initialization has zero or positive observed success is strongly associated with the post-RL result (point-biserial $r = 0.71$): zero-success initializations, all of which are control-arm tasks, end at 26.0\% on average, against 91.7\% for initializations with at least one observed success.
Among SynthDemo-SFT initializations on the Position axis, a weak one ($0 < \text{SFT} < 50\%$) ends at 97.0\% and a strong one ($\ge 50\%$) at 98.4\%, a 1.4-point difference against the ${\sim}91$-point gap between zero-success and weak initializations.
This correlation does not isolate the effect of coverage: task difficulty and the synthesis-plus-SFT intervention also vary, and SynthDemo-RL rates are near ceiling.

Table~\ref{tab:pertask} gives the per-task view: the control also drives 3 tasks with nonzero success before adaptation to 0\%, whereas SynthDemo-SFT loses none, and all tasks retain observed successes after PPO.

\begin{table}[t]
\centering\small
\caption{Per-task success (\%) on LIBERO-PRO (50 trials/task). Original: $\pi_{0.5}$-LIBERO before adaptation. Ctrl: direct PPO at matched RL compute. Ours: SynthDemo SFT, then PPO under the same RL compute as Ctrl (blue rows). Goal omits Task~3. Task-Goal Task~0 is excluded and not scored (ill-posed, see the footnote in Sec.~\ref{sec:exp:benchmarks}). Gray cells: the 27 tasks with 0\% Original success. Red cells: 0\% success. Bold: best result per task.}
\label{tab:pertask}
\resizebox{\columnwidth}{!}{%
\begin{tabular}{@{}llrrrrrrrrrr@{}}
\toprule
Suite & Method & 0 & 1 & 2 & 3 & 4 & 5 & 6 & 7 & 8 & 9 \\
\midrule
\multicolumn{12}{@{}l}{\textit{Position perturbations}} \\
Spatial & Original & 96 & 30 & \best{100} & \zerocell & 60 & 98 & \zerocell & \zerocell & \best{100} & \zerocell \\
Spatial & Ctrl PPO & \best{98} & 22 & \best{100} & \zerocell & 90 & 98 & \zerocell & \zerocell & \best{100} & \zerocell \\
Spatial & \cellcolor{blue!8}Ours (SFT) & 62 & 42 & 64 & \initzero{62} & 38 & 96 & \initzero{18} & \initzero{66} & 28 & \initzero{74} \\
Spatial & \cellcolor{blue!8}Ours (+RL) & 92 & \best{100} & 98 & \initzero{\best{96}} & \best{94} & \best{100} & \initzero{\best{96}} & \initzero{\best{100}} & 92 & \initzero{\best{96}} \\
\cmidrule(lr){1-12}
Object & Original & 30 & \best{98} & \zerocell & 2 & \zerocell & \zerocell & 34 & \zerocell & \zerocell & 4 \\
Object & Ctrl PPO & \best{100} & \best{98} & \zerocell & \zerocell & \zerocell & \zerocell & \best{100} & \zerocell & \initzero{4} & 34 \\
Object & \cellcolor{blue!8}Ours (SFT) & 96 & 32 & \initzero{18} & 96 & \initzero{94} & \initzero{94} & 28 & \initzero{96} & \initzero{24} & 6 \\
Object & \cellcolor{blue!8}Ours (+RL) & \best{100} & 92 & \initzero{\best{100}} & \best{100} & \initzero{\best{100}} & \initzero{\best{100}} & \best{100} & \initzero{\best{100}} & \initzero{\best{98}} & \best{100} \\
\cmidrule(lr){1-12}
Goal & Original & \zerocell & 66 & \zerocell & -- & \zerocell & \zerocell & \zerocell & 88 & 68 & \zerocell \\
Goal & Ctrl PPO & \zerocell & 90 & \zerocell & -- & \zerocell & \zerocell & \zerocell & \best{100} & 98 & \initzero{84} \\
Goal & \cellcolor{blue!8}Ours (SFT) & \initzero{78} & 8 & \initzero{16} & -- & \initzero{94} & \initzero{6} & \initzero{78} & 90 & 92 & \initzero{74} \\
Goal & \cellcolor{blue!8}Ours (+RL) & \initzero{\best{96}} & \best{98} & \initzero{\best{100}} & -- & \initzero{\best{100}} & \initzero{\best{94}} & \initzero{\best{94}} & \best{100} & \best{100} & \initzero{\best{100}} \\
\midrule
\multicolumn{12}{@{}l}{\textit{Task perturbations}} \\
Spatial & Original & \zerocell & 90 & \zerocell & 12 & \zerocell & \best{98} & 92 & 96 & 24 & \best{100} \\
Spatial & Ctrl PPO & \initzero{2} & \best{100} & \zerocell & \best{100} & \initzero{\best{100}} & \best{98} & \best{96} & \best{98} & 98 & \best{100} \\
Spatial & \cellcolor{blue!8}Ours (SFT) & \initzero{68} & 60 & \initzero{42} & 46 & \initzero{46} & 64 & 20 & 58 & 72 & 34 \\
Spatial & \cellcolor{blue!8}Ours (+RL) & \initzero{\best{96}} & \best{100} & \initzero{\best{94}} & 96 & \initzero{98} & 88 & 92 & 96 & \best{100} & 90 \\
\cmidrule(lr){1-12}
Object & Original & \zerocell & 96 & \zerocell & \zerocell & \zerocell & \zerocell & \zerocell & 12 & \zerocell & \zerocell \\
Object & Ctrl PPO & \initzero{\best{100}} & \best{100} & \initzero{36} & \initzero{\best{100}} & \zerocell & \zerocell & \initzero{88} & \zerocell & \zerocell & \initzero{\best{100}} \\
Object & \cellcolor{blue!8}Ours (SFT) & \initzero{12} & \best{100} & \initzero{76} & \initzero{10} & \initzero{\best{100}} & \initzero{90} & \initzero{72} & 32 & \initzero{92} & \initzero{54} \\
Object & \cellcolor{blue!8}Ours (+RL) & \initzero{\best{100}} & \best{100} & \initzero{\best{100}} & \initzero{\best{100}} & \initzero{\best{100}} & \initzero{\best{100}} & \initzero{\best{100}} & \best{100} & \initzero{\best{98}} & \initzero{98} \\
\cmidrule(lr){1-12}
Goal & Original & -- & 20 & 24 & -- & 8 & 2 & 24 & 90 & 48 & \zerocell \\
Goal & Ctrl PPO & -- & \zerocell & 88 & -- & 22 & \best{100} & 92 & \best{100} & \best{98} & \initzero{88} \\
Goal & \cellcolor{blue!8}Ours (SFT) & -- & 26 & 22 & -- & 4 & 98 & 22 & \best{100} & 20 & \initzero{50} \\
Goal & \cellcolor{blue!8}Ours (+RL) & -- & \best{100} & \best{94} & -- & \best{100} & \best{100} & \best{100} & \best{100} & 80 & \initzero{\best{100}} \\
\bottomrule
\end{tabular}}
\end{table}

\textbf{SFT shows forgetting on previously solved target tasks.}
Where $\pi_{0.5}$-LIBERO already solved some tasks, distilling perturbation-specific data costs performance on those tasks: on Task-Spatial, five of ten tasks drop after SFT (e.g., $92 \to 20$, $100 \to 34$).
RL largely closes these regressions: after RL, all five are back within 10 points of their original values (task~6: $92 \to 20 \to 92$).
On Position-Spatial, the four tasks the original policy solved at 96 to 100\% end within 8 points.

\textbf{Replay does not explain the gain.}
To test original-LIBERO replay, we compared Position-axis SFT on 50 synthetic episodes per task against the same recipe with 20 original episodes per task added.
Mean success was 57.7\% without replay and 51.5\% with, so replay did not improve target adaptation in this comparison.

\subsection{RoboTwin~2.0}
\label{sec:exp:robotwin}

To test the pipeline beyond LIBERO's simulator and embodiment, we apply it to four single-arm tasks with the Agilex Piper arm in RoboTwin~2.0.
Table~\ref{tab:robotwin_main} reports \texttt{success\_once} over 100 held-out trials per task.
SynthDemo-RL improves the SFT average from 55.5\% to 74.3\%, numerically close to SimpleVLA-RL.
The gain is largest on \texttt{beat\_hammer}, where SynthDemo-SFT is weakest (32\%), but the student remains below its teacher on two tasks.
For \texttt{place\_cup}, RL did not improve the strong SFT policy.

\begin{table}[t]
    \centering\small
    \caption{Success rates (\%) on four RoboTwin~2.0 single-arm tasks. Rows (a)--(c) use 50 trajectories per task synthesized by our teacher. Rows (e, e') use an OpenVLA-OFT backbone trained on RoboTwin's expert demonstrations (1{,}000 per task for SimpleVLA-RL), with results quoted from~\cite{simplevla-rl}. The comparison is contextual, not controlled.}
    \label{tab:robotwin_main}
    \resizebox{\columnwidth}{!}{%
    \begin{tabular}{@{}lccccc@{}}
        \toprule
        Method & \rotatebox{60}{\footnotesize place\_cup} & \rotatebox{60}{\footnotesize move\_can} & \rotatebox{60}{\footnotesize place\_a2b} & \rotatebox{60}{\footnotesize beat\_hammer} & Avg. \\
        \midrule
        \multicolumn{6}{@{}l}{\textit{Zero human demonstrations (ours)}} \\
        (a) SynthDemo-Teacher        & 94.0 & 57.0 & 86.0 & 80.0 & 79.3 \\
        (b) SynthDemo-SFT            & 96.0 & 38.0 & 56.0 & 32.0 & 55.5 \\
        (c) \textbf{SynthDemo-RL}    & \textbf{96.0} & 60.0 & \textbf{66.0} & 75.0 & \textbf{74.3} \\
        \midrule
        \multicolumn{6}{@{}l}{\textit{Baselines (RoboTwin expert-generated demonstrations)}} \\
        (e) OpenVLA-OFT~\cite{openvla-oft}         & 77.3 & 28.1 & 37.5 & 28.1 & 42.8 \\
        (e') $+$SimpleVLA-RL (w/ demo)~\cite{simplevla-rl} & 94.2 & \textbf{61.2} & 45.3 & \textbf{87.5} & 72.1 \\
        \bottomrule
    \end{tabular}}
\end{table}

\subsection{Physical Executability on a Real Robot}
\label{sec:exp:real}

\begin{figure}[t]
\centering
\includegraphics[width=\columnwidth]{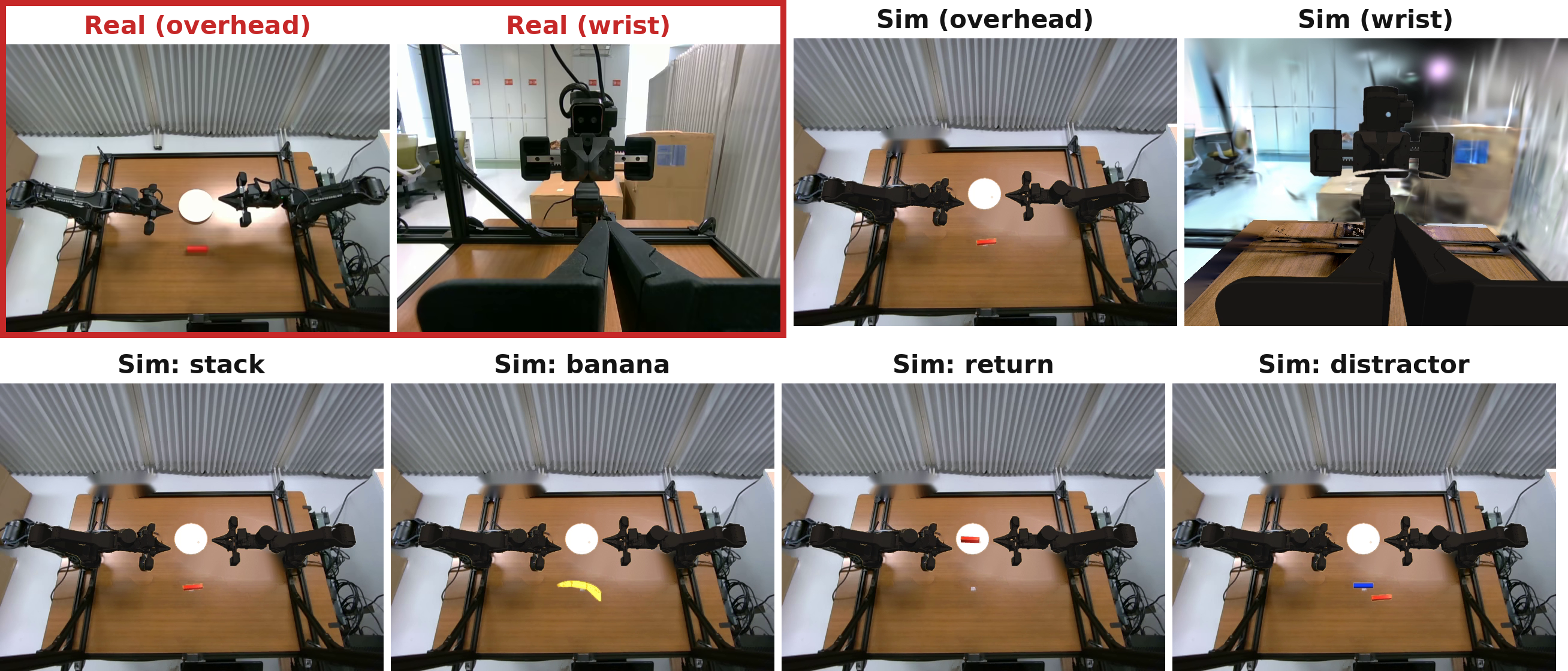}\\[3pt]
\resizebox{\columnwidth}{!}{%
\begin{tabular}{@{}lcccc@{}}
    \toprule
    Method & \rotatebox{60}{\footnotesize stack} & \rotatebox{60}{\footnotesize banana} & \rotatebox{60}{\footnotesize return} & \rotatebox{60}{\footnotesize distractor}  \\
    \midrule
    Human-demo reference, $\pi_{0.5}$ SFT, closed-loop & 50.0 & 0.0 & 0.0 & 0.0  \\
    \textbf{SynthDemo-RL}, no human demos, open-loop & 100.0 & 100.0 & 100.0 & 100.0  \\
    \bottomrule
\end{tabular}}
\caption{
\textbf{Physical executability on a real robot.}
Top: real camera views (red box) and the twin's hybrid renders of the same views.
Middle: the four conditions in the twin.
Bottom: hardware success (\%) over 20 trials per condition (target object released and at rest on its goal region). Ours executes twin rollouts open-loop, the reference runs closed-loop, so the rows are not a controlled comparison. Human demonstrations exist for \texttt{stack} only.}
\label{fig:real}
\end{figure}

We assess the physical executability of trajectories generated by a twin-trained SynthDemo-RL policy through open-loop execution on hardware; closed-loop sim-to-real transfer is outside our scope (Sec.~\ref{sec:limitations}).
The robot is a Trossen WidowX~AI stationary station (right arm, 30\,Hz, overhead and wrist RGB) with a MuJoCo twin of its workspace.
Using the same teacher and SFT-plus-PPO recipe, we train the $\pi_{0.5}$ student (LoRA adapters) entirely in the twin, with no fine-tuning on real-robot data.
The twin's renders composite the simulated arm and objects over the real camera backgrounds and a 3D~Gaussian~Splatting reconstruction of the room.
In preliminary closed-loop hardware trials the same checkpoint did not succeed, and an observation-sensitivity analysis suggests that the rendered appearance of the foreground (arm and plate) contributes to this gap.
We therefore test trajectory executability with open-loop execution, which uses no real-camera observations for policy feedback.
The policy runs closed-loop \emph{in the twin} from a matched initial state, and the resulting action sequence is executed on the robot without replanning, with objects placed by hand at the twin's nominal positions.
Four conditions share one workspace and one success criterion (Fig.~\ref{fig:real}): \texttt{stack} (the red block onto the plate), \texttt{banana} (a novel object), \texttt{return} (the reversed placement goal), and \texttt{distractor} (a blue block on the usual pick spot while the instruction names the red one).
Human teleoperation demonstrations exist for \texttt{stack} only, so the other three are the demonstration-free regime of this paper, now on hardware.
Open-loop execution succeeds in all 20 trials per condition. As context, the human-demonstration reference (trained on real images and run closed-loop) reaches 50\% on \texttt{stack} and 0\% on the three conditions for which it has no demonstrations.
The rows differ in training inputs and execution mode, so the table documents executability rather than a controlled comparison.

\subsection{Analysis}
\label{sec:exp:analysis}
\label{sec:exp:whysft}

The synthetic training set retains only successful episodes and matches the human demonstrations in policy, training budget, episode count, and initial-state provenance.
Even so, SFT reaches a three-suite average of 56.7\%, versus 97.7\% for human demonstrations (Table~\ref{tab:libero_main}).

\begin{table}[t]
    \centering\small
    \caption{\textbf{How far the synthetic demonstrations are from human ones, and what closing that distance buys.} Per-episode medians on LIBERO-Spatial for the original synthetic data (S), the humanized-teacher regeneration (S$^\dagger$), and human demonstrations (H). Cliff's $\delta$ compares S against H ($n = 864$ episodes per source; Mann--Whitney $p < 10^{-48}$ for every row except within-task DTW diversity, $p = 0.28$). The frozen-base loss scores each dataset under the unmodified $\pi_{0.5}$ base checkpoint with identical normalization: a descriptive measure of mismatch with the frozen model.}
    \label{tab:distshift}
    \resizebox{\columnwidth}{!}{%
    \begin{tabular}{@{}lrrrr@{}}
        \toprule
        Measure & S & S$^\dagger$ & H & $\delta$ \\
        \midrule
        Action saturation (frac.\ frames) & 0.158 & 0.000 & 0.027 & 1.00 \\
        ~~vertical axis only              & 0.498 & --    & 0.094 & 1.00 \\
        Jerk norm                         & 0.125 & 0.087 & 0.069 & 0.997 \\
        Velocity norm                     & 0.118 & --    & 0.099 & 0.58 \\
        Frozen $\pi_{0.5}$ base FM loss   & 0.123 & 0.090 & 0.093 & 0.89 \\
        Within-task DTW diversity         & 0.199 & --    & 0.200 & $-0.02$ \\
        Episode length (steps)            & 124   & 153   & 123   & -- \\
        \midrule
        \textbf{SFT success (\%)}         & \textbf{55.3}$\pm$5.1 & \textbf{60.2}$\pm$1.5 & \textbf{97.8} & -- \\
        \bottomrule
    \end{tabular}}
\end{table}

\begin{figure}[t]
\centering
\includegraphics[width=\columnwidth]{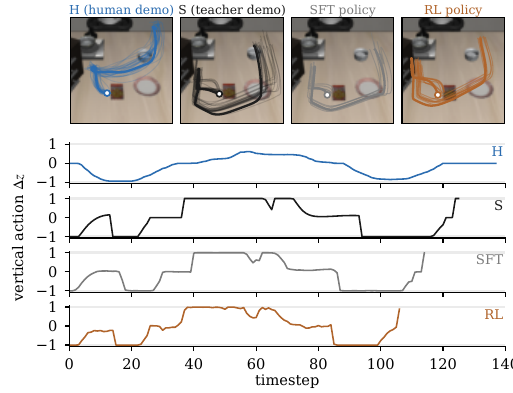}
\caption{
\textbf{From teacher signature to policy behavior.}
Top: end-effector paths on one LIBERO-Spatial task: human demonstrations (H), teacher demonstrations (S), and successful rollouts before (SFT) and after RL, overlaid thin. The episode closest to each source's median detour ratio (path length over start--end displacement) is bold. SFT and RL share the initial state, and all panels share one crop.
Bottom: vertical action of the bold episodes; gray bands mark saturation ($|a| \ge 0.9375$).
RL straightens the geometry (median detour ratio $3.7 \to 3.2$ vs 2.5 for human) but keeps the teacher's saturation signature: median saturation 0.126 (SFT) and 0.133 (RL) against 0.158 (S) and 0.027 (H).
}
\label{fig:traj}
\end{figure}

\textbf{Synthetic and human demonstrations differ in motion statistics and frozen-base loss.}
Table~\ref{tab:distshift} quantifies the difference.
Kinematic statistics separate the two sources completely: saturated-action frames and jerk norm reach $\delta \approx 1$, driven by the vertical axis.
This servo signature is directly visible in Fig.~\ref{fig:traj}, and it is not simply speed (velocity separates far more weakly, $\delta = 0.58$).
Under the frozen $\pi_{0.5}$ base checkpoint, flow-matching loss is systematically higher for synthetic episodes ($\delta = 0.89$ on Spatial, $0.99$ on Object, and $0.52$ on Goal), and saturation and jerk separate with $\delta \ge 0.99$ on all three suites. This indicates greater prediction error under the flow-matching objective, not a direct estimate of trajectory likelihood.
Robometer-4B~\cite{robometer}, a pretrained vision-language reward model that estimates per-frame progress from video and instructions, assigns lower progress monotonicity to synthetic episodes on Spatial and Goal (both $p < 0.02$), with similar final predicted progress.
On Spatial, within-task DTW diversity shows little observed difference between synthetic and human demonstrations ($\delta = -0.02$, $p = 0.28$), which does not point to a diversity deficit.

\textbf{The SFT gap persists after humanizing the teacher.}
Constraining the teacher's servo velocity and rotation limits to human-measured levels and regenerating the dataset (a two-line generator-configuration change) gives column S$^\dagger$: saturation vanishes and the frozen-base loss falls slightly below the human value.
The outcome moves far less: averaged over three seeds, regenerated-data SFT improves success from 55.3\% to 60.2\%, closing 4.9 of the 42.5-point Spatial gap to human-demonstration SFT (97.8\%).
Eight of ten tasks improve, and the regenerated episodes are 24\% longer.
This improvement is consistent with teacher-induced motion artifacts contributing to the SFT gap, but the regeneration does not isolate their effect or explain the remaining gap.
Trajectory shape and grasp semantics remain possible contributors; the Object-suite gap concentrates on three box-shaped objects that fit the enforced top-down grasp poorly.
Under the fixed training schedule RL raises SynthDemo-SFT from 55.3\% to 96.2\% on Spatial (Table~\ref{tab:libero_main}).
Fig.~\ref{fig:traj} illustrates accompanying changes in policy behavior.
SFT retains the teacher's saturation signature, while RL changes the trajectory geometry: the detour ratio closes 41\% of the SFT-to-human gap, while the per-frame $\Delta z$ distribution moves \emph{away} from the human profile on all ten tasks ($W_1$ distance).
The 41-point gain therefore occurs without making the measured vertical-action distribution more human-like.

\section{Limitations and Future Work}
\label{sec:limitations}

Our claim is human-demonstration-free adaptation with a simulator in the loop, not simulator-free learning: the teacher consumes ground-truth simulator state, and RL requires an environment to roll out in, so a target-domain simulation (as in Sec.~\ref{sec:exp:real}) is a prerequisite.
Replacing the teacher's privileged inputs with learned perception, or the RL simulator with a learned world model~\cite{vla-rft}, are untested extensions.
SFT without rehearsal can reduce success on previously solved target tasks, although subsequent RL largely recovers these losses (Sec.~\ref{sec:exp:novel}).
The teacher executes one primitive per task.
Multi-step manipulation (e.g., LIBERO-Long) would require sub-goal decomposition and sequencing of the existing primitives, which we have not implemented or evaluated.
Closed-loop sim-to-real transfer remains future work (Sec.~\ref{sec:exp:real}).
The LLM tuner relies on a proprietary API.

\section{Conclusion}
\label{sec:conclusion}

We presented SynthDemo-RL, a framework that adapts VLA manipulation policies to new task distributions without new human teleoperation demonstrations, and studied reward coverage as a diagnostic of the distilled initialization.
On the 57 scored perturbed LIBERO-PRO tasks, SynthDemo-SFT rescues all 27 tasks with no observed success before adaptation.
Subsequent PPO raises success to 97.8\% (Position) and 97.1\% (Task), while the PPO-matched sparse-reward control rescues 10 of the 27 and leaves 17 at 0\%.
The pipeline reaches 96.0\% on standard LIBERO without collecting new human demonstrations and carries over to RoboTwin~2.0. Its twin-trained trajectories execute open-loop on a physical robot.
These findings suggest that demonstration synthesis should prioritize reward coverage, at least one successful trajectory on every task, and leave raising the average success rate to sparse-reward RL.
\section*{Acknowledgments}
OpenAI tools (GPT-5.5 and Codex) assisted with automated teacher tuning and manuscript refinement.
\bibliographystyle{IEEEtran}
\bibliography{references,references_extra}

\end{document}